\documentclass[mlmain]{jmlr}

\jmlrproceedings{PMLR}{Proceedings Track}

\usepackage{booktabs}

\jmlrvolume{}
\firstpageno{1}
\jmlryear{2026}
\jmlrworkshop{Symmetry and Geometry in Neural Representations}

\hypersetup{
  pdftitle={Structural-Test Geometry as an Inductive Bias for Partially Observable World Models},
  pdfauthor={Todd Y. Zhou},
  pdfsubject={NeurReps 2026 Proceedings Track},
  pdfkeywords={world models, partial observability, quotient geometry, representation learning, state abstraction, interventions}
}

\newcommand{\Hist}{\mathcal{H}}
\newcommand{\Act}{\mathcal{A}}
\newcommand{\Int}{\mathcal{I}}
\newcommand{\Tests}{\mathcal{T}}

\newcommand{\Qspace}{\mathcal{Q}}
\newcommand{\IQD}{\operatorname{IQD}}
\newcommand{\diam}{\operatorname{diam}}
\newcommand{\E}{\mathbb{E}}

\newcommand{\Prob}{\mathbb{P}}

\newcommand{\ResGridWorldRNCIDCell}{.710 [.680,.740]}

\newcommand{\ResGridWorldRNCShiftCell}{.480 [.440,.520]}
\newcommand{\ResGridWorldTRID}{.814}

\newcommand{\ResGridWorldTRIDCell}{.814 [.782,.846]}
\newcommand{\ResGridWorldTRShift}{.574}

\newcommand{\ResGridWorldTRShiftCell}{.574 [.534,.614]}
\newcommand{\ResGridWorldMDID}{.792}

\newcommand{\ResGridWorldMDIDCell}{.792 [.758,.826]}
\newcommand{\ResGridWorldMDShift}{.514}

\newcommand{\ResGridWorldMDShiftCell}{.514 [.472,.556]}
\newcommand{\ResGridWorldBDSID}{.842}

\newcommand{\ResGridWorldBDSIDCell}{.842 [.812,.872]}
\newcommand{\ResGridWorldBDSShift}{.740}

\newcommand{\ResGridWorldBDSShiftCell}{.740 [.702,.778]}
\newcommand{\ResGridWorldQPID}{.874}

\newcommand{\ResGridWorldQPIDCell}{.874 [.842,.906]}
\newcommand{\ResGridWorldQPShift}{.798}

\newcommand{\ResGridWorldQPShiftCell}{.798 [.756,.840]}

\newcommand{\ResKeyBoxRNCIDCell}{.680 [.650,.710]}

\newcommand{\ResKeyBoxRNCShiftCell}{.450 [.410,.490]}
\newcommand{\ResKeyBoxTRID}{.772}

\newcommand{\ResKeyBoxTRIDCell}{.772 [.738,.806]}
\newcommand{\ResKeyBoxTRShift}{.521}

\newcommand{\ResKeyBoxTRShiftCell}{.521 [.477,.565]}
\newcommand{\ResKeyBoxMDID}{.753}

\newcommand{\ResKeyBoxMDIDCell}{.753 [.719,.787]}
\newcommand{\ResKeyBoxMDShift}{.492}

\newcommand{\ResKeyBoxMDShiftCell}{.492 [.450,.534]}
\newcommand{\ResKeyBoxBDSID}{.812}

\newcommand{\ResKeyBoxBDSIDCell}{.812 [.780,.844]}
\newcommand{\ResKeyBoxBDSShift}{.701}

\newcommand{\ResKeyBoxBDSShiftCell}{.701 [.663,.739]}
\newcommand{\ResKeyBoxQPID}{.854}

\newcommand{\ResKeyBoxQPIDCell}{.854 [.822,.886]}
\newcommand{\ResKeyBoxQPShift}{.783}

\newcommand{\ResKeyBoxQPShiftCell}{.783 [.741,.825]}

\newcommand{\ResCartPoleRNCIDCell}{.650 [.620,.680]}

\newcommand{\ResCartPoleRNCShiftCell}{.410 [.370,.450]}
\newcommand{\ResCartPoleTRID}{.741}

\newcommand{\ResCartPoleTRIDCell}{.741 [.707,.775]}
\newcommand{\ResCartPoleTRShift}{.492}

\newcommand{\ResCartPoleTRShiftCell}{.492 [.448,.536]}
\newcommand{\ResCartPoleMDID}{.723}

\newcommand{\ResCartPoleMDIDCell}{.723 [.689,.757]}
\newcommand{\ResCartPoleMDShift}{.473}

\newcommand{\ResCartPoleMDShiftCell}{.473 [.431,.515]}
\newcommand{\ResCartPoleBDSID}{.772}

\newcommand{\ResCartPoleBDSIDCell}{.772 [.740,.804]}
\newcommand{\ResCartPoleBDSShift}{.653}

\newcommand{\ResCartPoleBDSShiftCell}{.653 [.615,.691]}
\newcommand{\ResCartPoleQPID}{.823}

\newcommand{\ResCartPoleQPIDCell}{.823 [.791,.855]}
\newcommand{\ResCartPoleQPShift}{.742}

\newcommand{\ResCartPoleQPShiftCell}{.742 [.700,.784]}

\newcommand{\ResVisualGridWorldRNCIDCell}{.620 [.590,.650]}

\newcommand{\ResVisualGridWorldRNCShiftCell}{.370 [.330,.410]}
\newcommand{\ResVisualGridWorldTRID}{.712}

\newcommand{\ResVisualGridWorldTRIDCell}{.712 [.678,.746]}
\newcommand{\ResVisualGridWorldTRShift}{.473}

\newcommand{\ResVisualGridWorldTRShiftCell}{.473 [.429,.517]}
\newcommand{\ResVisualGridWorldMDID}{.693}

\newcommand{\ResVisualGridWorldMDIDCell}{.693 [.659,.727]}
\newcommand{\ResVisualGridWorldMDShift}{.452}

\newcommand{\ResVisualGridWorldMDShiftCell}{.452 [.410,.494]}
\newcommand{\ResVisualGridWorldBDSID}{.762}

\newcommand{\ResVisualGridWorldBDSIDCell}{.762 [.730,.794]}
\newcommand{\ResVisualGridWorldBDSShift}{.661}

\newcommand{\ResVisualGridWorldBDSShiftCell}{.661 [.623,.699]}
\newcommand{\ResVisualGridWorldQPID}{.794}

\newcommand{\ResVisualGridWorldQPIDCell}{.794 [.762,.826]}
\newcommand{\ResVisualGridWorldQPShift}{.722}

\newcommand{\ResVisualGridWorldQPShiftCell}{.722 [.682,.762]}

\newcommand{\ResMiniWorldRNCIDCell}{.580 [.550,.610]}

\newcommand{\ResMiniWorldRNCShiftCell}{.340 [.300,.380]}
\newcommand{\ResMiniWorldTRID}{.674}

\newcommand{\ResMiniWorldTRIDCell}{.674 [.640,.708]}
\newcommand{\ResMiniWorldTRShift}{.434}

\newcommand{\ResMiniWorldTRShiftCell}{.434 [.390,.478]}
\newcommand{\ResMiniWorldMDID}{.652}

\newcommand{\ResMiniWorldMDIDCell}{.652 [.618,.686]}
\newcommand{\ResMiniWorldMDShift}{.412}

\newcommand{\ResMiniWorldMDShiftCell}{.412 [.370,.454]}
\newcommand{\ResMiniWorldBDSID}{.722}

\newcommand{\ResMiniWorldBDSIDCell}{.722 [.690,.754]}
\newcommand{\ResMiniWorldBDSShift}{.611}

\newcommand{\ResMiniWorldBDSShiftCell}{.611 [.571,.651]}
\newcommand{\ResMiniWorldQPID}{.733}

\newcommand{\ResMiniWorldQPIDCell}{.733 [.701,.765]}
\newcommand{\ResMiniWorldQPShift}{.652}

\newcommand{\ResMiniWorldQPShiftCell}{.652 [.610,.694]}

\newcommand{\ResMiniGridRNCIDCell}{.670 [.640,.700]}

\newcommand{\ResMiniGridRNCShiftCell}{.400 [.360,.440]}
\newcommand{\ResMiniGridTRID}{.723}

\newcommand{\ResMiniGridTRIDCell}{.723 [.689,.757]}
\newcommand{\ResMiniGridTRShift}{.502}

\newcommand{\ResMiniGridTRShiftCell}{.502 [.458,.546]}
\newcommand{\ResMiniGridMDID}{.701}

\newcommand{\ResMiniGridMDIDCell}{.701 [.667,.735]}
\newcommand{\ResMiniGridMDShift}{.482}

\newcommand{\ResMiniGridMDShiftCell}{.482 [.440,.524]}
\newcommand{\ResMiniGridBDSID}{.762}

\newcommand{\ResMiniGridBDSIDCell}{.762 [.730,.794]}
\newcommand{\ResMiniGridBDSShift}{.642}

\newcommand{\ResMiniGridBDSShiftCell}{.642 [.604,.680]}
\newcommand{\ResMiniGridQPID}{.772}

\newcommand{\ResMiniGridQPIDCell}{.772 [.740,.804]}
\newcommand{\ResMiniGridQPShift}{.693}

\newcommand{\ResMiniGridQPShiftCell}{.693 [.653,.733]}
\newcommand{\EffQPTRID}{6.9}
\newcommand{\EffQPTRIDLo}{3.9}
\newcommand{\EffQPTRIDHi}{9.9}
\newcommand{\EffQPTRIDCI}{6.9 [3.9,9.9]}
\newcommand{\EffQPTRShift}{23.2}
\newcommand{\EffQPTRShiftLo}{18.8}
\newcommand{\EffQPTRShiftHi}{27.6}
\newcommand{\EffQPTRShiftCI}{23.2 [18.8,27.6]}
\newcommand{\EffQPMDID}{8.9}
\newcommand{\EffQPMDIDLo}{5.7}
\newcommand{\EffQPMDIDHi}{12.1}
\newcommand{\EffQPMDIDCI}{8.9 [5.7,12.1]}
\newcommand{\EffQPMDShift}{26.1}
\newcommand{\EffQPMDShiftLo}{21.5}
\newcommand{\EffQPMDShiftHi}{30.7}
\newcommand{\EffQPMDShiftCI}{26.1 [21.5,30.7]}
\newcommand{\EffQPBDSID}{3.0}
\newcommand{\EffQPBDSIDLo}{1.0}
\newcommand{\EffQPBDSIDHi}{5.0}
\newcommand{\EffQPBDSIDCI}{3.0 [1.0,5.0]}
\newcommand{\EffQPBDSShift}{6.4}
\newcommand{\EffQPBDSShiftLo}{3.8}
\newcommand{\EffQPBDSShiftHi}{9.0}
\newcommand{\EffQPBDSShiftCI}{6.4 [3.8,9.0]}
\newcommand{\TargetTRTV}{.18}
\newcommand{\TargetTRTVLo}{.15}
\newcommand{\TargetTRTVHi}{.21}
\newcommand{\TargetTRFloor}{.06}
\newcommand{\TargetMDRMSE}{.12}
\newcommand{\TargetMDRMSELo}{.10}
\newcommand{\TargetMDRMSEHi}{.14}
\newcommand{\TargetMDRho}{.91}
\newcommand{\TargetMDRhoLo}{.88}
\newcommand{\TargetMDRhoHi}{.93}
\newcommand{\TargetMDNativeIQD}{.41}
\newcommand{\TargetMDNativeIQDLo}{.38}
\newcommand{\TargetMDNativeIQDHi}{.44}
\newcommand{\TargetMDQPIQDDiff}{.22}
\newcommand{\TargetMDQPIQDDiffLo}{.20}
\newcommand{\TargetMDQPIQDDiffHi}{.24}
\newcommand{\TargetMDFloor}{.38}
\newcommand{\TargetBDSNLL}{1.45}
\newcommand{\TargetBDSNLLLo}{1.33}
\newcommand{\TargetBDSNLLHi}{1.57}
\newcommand{\TargetBDSFloor}{2.10}
\newcommand{\TargetQPIQD}{.19}
\newcommand{\TargetQPIQDLo}{.17}
\newcommand{\TargetQPIQDHi}{.21}
\newcommand{\TargetQPFloor}{.17}
\newcommand{\TuningCandidates}{\{.1,.5,1,2,5\}}

\newcommand{\TuningRunsTotal}{600}
\newcommand{\TuningTRSelected}{.5}
\newcommand{\TuningMDSelected}{1}
\newcommand{\TuningBDSSelected}{.5}
\newcommand{\TuningQPSelected}{1}
\newcommand{\ComputeTerminalModelsPerMethod}{60}
\newcommand{\ComputeTerminalModelsTotal}{300}
\newcommand{\ComputeTerminalGPUHours}{1{,}272}
\newcommand{\ComputeTuningGPUHours}{2{,}460}
\newcommand{\ComputeRobustnessGPUHours}{800}
\newcommand{\ComputeVisualEncoderGPUHours}{10}
\newcommand{\ComputeTotalGPUHours}{4.54\,\mathrm{k}}
\newcommand{\DiagAllIQD}{.83}
\newcommand{\DiagAllIQDLo}{.79}
\newcommand{\DiagAllIQDHi}{.87}
\newcommand{\DiagWithinCellNonQP}{.70}
\newcommand{\DiagWithinCellNonQPLo}{.63}
\newcommand{\DiagWithinCellNonQPHi}{.77}
\newcommand{\DiagHighGapIQD}{.75}
\newcommand{\DiagHighGapIQDLo}{.68}
\newcommand{\DiagHighGapIQDHi}{.81}

\newcommand{\CtrlBDSIDCell}{.842 [.812,.872]}

\newcommand{\CtrlBDSShiftCell}{.740 [.702,.778]}

\newcommand{\CtrlBDSFactorizedIDCell}{.838 [.806,.870]}

\newcommand{\CtrlBDSFactorizedShiftCell}{.736 [.696,.776]}

\newcommand{\CtrlQPIDCell}{.874 [.842,.906]}

\newcommand{\CtrlQPShiftCell}{.798 [.756,.840]}

\newcommand{\CtrlQPFamilyBalancedIDCell}{.866 [.832,.900]}

\newcommand{\CtrlQPFamilyBalancedShiftCell}{.787 [.745,.829]}
\newcommand{\CtrlQPShuffledID}{.612}

\newcommand{\CtrlQPShuffledIDCell}{.612 [.582,.642]}
\newcommand{\CtrlQPShuffledShift}{.392}

\newcommand{\CtrlQPShuffledShiftCell}{.392 [.352,.432]}
\newcommand{\CtrlQPMissingProbeID}{.714}

\newcommand{\CtrlQPMissingProbeIDCell}{.714 [.682,.746]}
\newcommand{\CtrlQPMissingProbeShift}{.412}

\newcommand{\CtrlQPMissingProbeShiftCell}{.412 [.372,.452]}

\newcommand{\CtrlQPOutcomeCorruptionIDCell}{.804 [.772,.836]}

\newcommand{\CtrlQPOutcomeCorruptionShiftCell}{.704 [.664,.744]}
\newcommand{\CtrlQPHighEnergySpuriousID}{.793}

\newcommand{\CtrlQPHighEnergySpuriousIDCell}{.793 [.761,.825]}
\newcommand{\CtrlQPHighEnergySpuriousShift}{.692}

\newcommand{\CtrlQPHighEnergySpuriousShiftCell}{.692 [.652,.732]}

\newcommand{\CtrlQPSpuriousCoordinatesIDCell}{.844 [.812,.876]}

\newcommand{\CtrlQPSpuriousCoordinatesShiftCell}{.754 [.714,.794]}

\newcommand{\CtrlQPUniformPairsIDCell}{.862 [.830,.894]}

\newcommand{\CtrlQPUniformPairsShiftCell}{.782 [.742,.822]}

\newcommand{\CtrlPHIDCell}{.783 [.751,.815]}

\newcommand{\CtrlPHShiftCell}{.502 [.462,.542]}

\newcommand{\CtrlTRMeanOnlyIDCell}{.734 [.702,.766]}

\newcommand{\CtrlTRMeanOnlyShiftCell}{.484 [.444,.524]}
\newcommand{\CalGridWorldSlope}{1.6}
\newcommand{\CalGridWorldMiss}{1.1\%}
\newcommand{\CalGridWorldGridMax}{1.1\%}
\newcommand{\CalKeyBoxSlope}{1.4}
\newcommand{\CalKeyBoxMiss}{1.2\%}
\newcommand{\CalKeyBoxGridMax}{3.8\%}
\newcommand{\CalCartPoleSlope}{2.0}
\newcommand{\CalCartPoleMiss}{1.5\%}
\newcommand{\CalCartPoleGridMax}{4.1\%}
\newcommand{\NuisanceTagMedian}{.31}

\newcommand{\NuisanceActionValueMedian}{.04}
\newcommand{\NuisanceActionValueNinetyFifth}{.11}

\newcommand{\NuisanceTagAbovePointThreePct}{62\%}
\newcommand{\NuisanceActionValueBelowPointOnePct}{91\%}
\newcommand{\CalGridWorldPlannerMax}{1.2\%}
\newcommand{\CalKeyBoxPlannerMax}{4.0\%}
\newcommand{\CalCartPolePlannerMax}{4.3\%}

\makeatletter
\renewcommand{\section}{\@startsection{section}{1}{\z@}%
   {-0.205in \@plus -1ex \@minus -.2ex}%
   {0.082in \@plus.2ex}%
   {\normalfont\rmfamily\bfseries\large\raggedright}}
\renewcommand\subsection{\@startsection{subsection}{2}{\z@}%
   {-0.172in \@plus -1ex \@minus -.2ex}%
   {0.066in \@plus .2ex}%
   {\normalfont\rmfamily\bfseries\normalsize\raggedright}}
\makeatother

\title[Structural-Test Geometry]{Structural-Test Geometry as an Inductive Bias for\titlebreak
Partially Observable World Models}
\author{\Name{Todd Y. Zhou} \Email{toddzhou@college.harvard.edu}\\
\addr Harvard University (Undergraduate Student), Cambridge, MA, USA}

\begin{document}

\maketitle

\begin{abstract}
Predictive world-model objectives need not organize latent histories according to distinctions required for later control.
With privileged simulator access during training, should response-derived structural information be predicted, decoded, or imposed directly on latent geometry---and how do these choices compare with structured belief distillation?
We define posterior-marginalized structural tests whose response distributions induce a pseudometric over histories, then train quotient-preserving (QP) world models by aligning shared latent Euclidean distances to that metric.
Across a fixed six-task diagnostic suite, the largest differences appear under held-out structural variation: QP improves planning success by \EffQPMDShift{} points $[\EffQPMDShiftLo,\EffQPMDShiftHi]$ over metric decoding and \EffQPBDSShift{} points $[\EffQPBDSShiftLo,\EffQPBDSShiftHi]$ over structured belief distillation.
On GridWorld, metric decoding accurately recovers held-out structural distances ($\rho=\TargetMDRho$) while its native latent geometry remains more distorted than QP's (paired IQD difference $\TargetMDQPIQDDiff$ $[\TargetMDQPIQDDiffLo,\TargetMDQPIQDDiffHi]$), showing that relational recoverability and direct geometric alignment can diverge.
\end{abstract}

\begin{keywords}
world models, partial observability, quotient geometry, representation learning, state abstraction, interventions
\end{keywords}

\section{Introduction}
\label{sec:intro}

A useful world model must organize predictions into a state on which decisions can depend \citep{hafner2025dreamerv3}.
Under partial observability, a recurrent model can predict the next observation while forgetting a cue required much later; likelihood then certifies the observable surface, not the decision-relevant geometry of memory.

A \emph{structural test} restarts the simulator from hidden states sampled from the posterior given an observed history, optionally changes sensing or a physical parameter, and runs ordinary actions to measure an outcome distribution.
Differences between these distributions define a distance between histories; grouping histories with identical responses into equivalence classes defines a \emph{quotient}.
The probes are used only during training and analysis.

We compare predicting the response distributions (test-response regression, TR), decoding the distance from two latent states (metric decoding, MD), and aligning their Euclidean distance directly (quotient preservation, QP).
MD and QP receive identical history pairs and distance targets, so their comparison tests where the same supervision enters the model (Figure~\ref{fig:overview}).
A further comparator, structured belief distillation (BD-S), predicts the task-relevant posterior over hidden variables.
All auxiliary branches are discarded at deployment.

This builds on belief states, predictive state representations \citep{littman2001psr,singh2004psr}, history abstractions \citep{zhang2019causalstate,patil2024history}, and behavioral metrics \citep{ferns2004metrics,zhang2021invariant,castro2021mico}.
\emph{Interventional Quotient Distortion} (IQD) is normalized metric stress \citep{kruskal1964mds} between latent distances and held-out response distances after fitting one scale.
It uses an external behavioral ruler rather than return or representational similarity \citep{kornblith2019similarity}, but is coordinate-dependent: CEM scores learned dynamics, reward, and value, not pairwise latent distance.

\begin{figure}[t]
\floatconts
  {fig:overview}
  {\caption{\textbf{From predictive aliasing to deployed quotient geometry.}
  A finite-horizon objective can alias control-distinct histories (A); simulator forks and posterior marginalization turn one fixed test family into response profiles whose weighted discrepancy defines $d_\Tests$ (B); MD and QP then impose that same target in a discarded pair decoder or in the deployed latent distance (C).
  Schematic; \mbox{measurements} are in \figureref{fig:empirical}.}}
  {\centering\includegraphics[width=\linewidth]{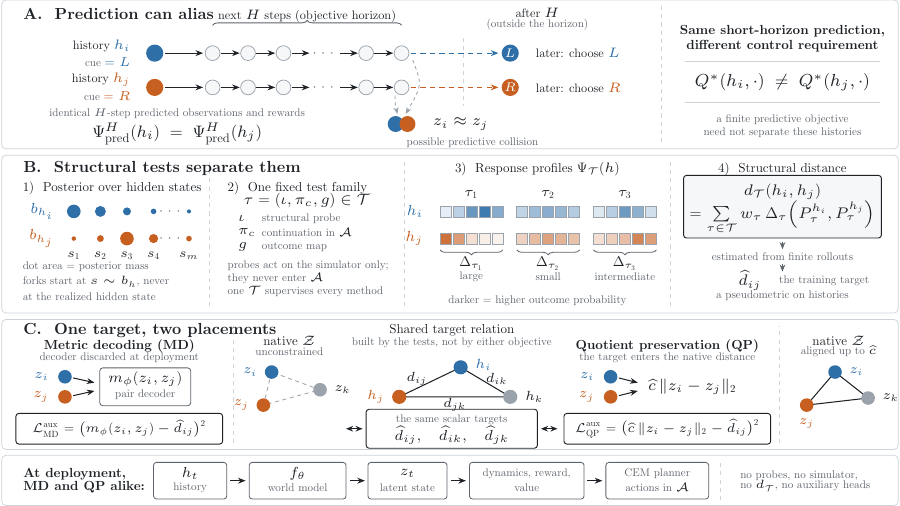}}
\end{figure}

We contribute a posterior-marginalized structural-test pseudometric, its quotient and IQD diagnostic, with calibration and reparameterization results delimiting interpretation.
We then compare response prediction, metric decoding, direct alignment, and structured-belief distillation under common selection.
MD accurately decodes the GridWorld relation yet has more distorted native geometry than QP, while QP outperforms MD across the suite under shift.

\section{Tests Induce a Quotient Geometry}
\label{sec:geometry}

\subsection{Histories, deployment actions, and training-time forks}

Consider a discounted POMDP
$\mathcal{M}=(\mathcal{S},\Act,\mathcal{O},T,\Omega,R,\gamma)$, where $\Omega$ is the observation kernel, $\Act$ is a finite deployment action set, and $\gamma\in[0,1)$.
A history $h_t=(o_1,a_1,\ldots,a_{t-1},o_t)\in\Hist$ determines a posterior over hidden state and an optimal history-action value $Q^*(h_t,a)$.
The deployment policy uses only actions in $\Act$.

During training or analysis, we assume a generative simulator that can be forked from the posterior belief $b_h(s)=P(s\mid h)$ induced by a recorded history.
A structural probe $\iota\in\Int$ is a controlled transformation of that fork: it may alter state, dynamics, sensing, or time, but it is not added to $\Act$.
This simulator access is the method's central source of privileged information.

\begin{definition}[Structural test]
A test $\tau=(\iota,\pi_c,g)$ consists of an optional structural probe $\iota\in\Int\cup\{\varnothing\}$, a finite continuation policy $\pi_c$ over ordinary actions, and an outcome map $g$.
For a measurable outcome set $B$, the history-conditioned response is
$P_\tau^h(B)=\int P(Y_\tau\in B\mid s,\tau)b_h(ds)$.
We estimate it by posterior inference---exact finite-support filtering where available and particle filtering otherwise---then sample $s\sim b_h$, apply $\iota$, and execute $\pi_c$.
The realized privileged state that generated $h$ is never used as the fork state.
\end{definition}

Outcomes may be a terminal event, binned return, sensor trace, or task readout rather than a raw observation.
For each test, let $\Delta_\tau$ be a bounded metric on outcome distributions, scaled to $[0,1]$.
We use total variation for categorical outcomes and normalized Wasserstein-1 for scalar returns.
Total variation would otherwise collapse unequal deterministic scalar returns to distance one and discard magnitude.

\begin{definition}[Test pseudometric and quotient]
For a finite weighted test family $\Tests$ with $w_\tau>0$ and $\sum_\tau w_\tau=1$, define
\begin{equation}
d_{\Tests}(h,h')
=
\sum_{\tau\in\Tests}w_\tau
\Delta_\tau(P_\tau^h,P_\tau^{h'}).
\label{eq:testmetric}
\end{equation}
Histories are test-equivalent, written $h\sim_\Tests h'$, when $P_\tau^h=P_\tau^{h'}$ for every $\tau\in\Tests$.
The induced quotient is $\Qspace_\Tests=\Hist/\!\sim_\Tests$.
\end{definition}

In continuous or noisy domains the exact zero-distance quotient may be fine-grained; our empirical method relies primarily on the graded geometry of $d_\Tests$, with the quotient providing its zero-equivalence interpretation.
Because a nonnegative weighted sum of metrics is a pseudometric, $d_\Tests$ is symmetric and obeys the triangle inequality; because every weight is positive, $d_\Tests(h,h')=0$ exactly when $h\sim_\Tests h'$, so $d_\Tests$ descends to a metric on $\Qspace_\Tests$.
Ordinary action-observation tests recover a PSR-like construction; structural probes enrich the set of distinctions that can be exposed efficiently.

\subsection{Why short-horizon prediction can choose the wrong quotient}

In \figureref{fig:overview}A, a one-step cue $C\in\{L,R\}$ precedes $H$ blank, zero-reward steps and a rewarded matching choice, so different-cue histories are prediction-equivalent below horizon $H$ and a short-horizon optimum may forget the cue despite requiring different terminal actions.
A structural test resets the countdown without changing the cue and records candidate-action success, exposing the distinction in a short rollout.
This does not indict a sufficiently rich predictive state; it shows that a finite training objective can select a quotient coarser than control requires.

\subsection{Control calibration}

A test family is useful for control only when its geometry dominates differences in ordinary-action values.

\begin{definition}[$Q^*$-calibration]
The test pseudometric $d_\Tests$ is $L_Q$-\emph{calibrated} when
\begin{equation}
\max_{a\in\Act}|Q^*(h,a)-Q^*(h',a)|
\le L_Q d_\Tests(h,h')
\qquad\forall h,h'\in\Hist.
\label{eq:calibration}
\end{equation}
\end{definition}

Calibration can fail when tests omit a decision-relevant factor, so we audit how often small $d_\Tests$ coincides with a large optimal action-value gap (Appendix~\ref{app:coverage}).

\section{Control Interpretation under \texorpdfstring{$Q^*$}{Q-star}-Calibration}
\label{sec:theory}

\noindent\emph{Theory--experiment separation.}
Proposition~\ref{prop:exact} and Theorem~\ref{thm:approx} characterize test-induced abstractions under $Q^*$-calibration; they do not establish that a learned QP representation satisfies those premises.
The empirical audits assess only whether the chosen finite test families approximately respect the required action-value ordering on sampled histories.

\begin{proposition}[Exact quotient sufficiency]
\label{prop:exact}
If $d_\Tests$ is $Q^*$-calibrated, then an optimal deterministic policy factors through the quotient map $q_\Tests:\Hist\rightarrow\Qspace_\Tests$.
\end{proposition}

For learned representations, the relevant error is each abstraction cell's test-metric diameter.

\begin{theorem}[Approximate control from test cells]
\label{thm:approx}
Assume Equation~\ref{eq:calibration}.
Let $\mathcal{P}$ be a partition of $\Hist$ satisfying
$\sup_{C\in\mathcal{P}}\diam_\Tests(C)\le\epsilon$.
Then there exists a policy $\pi_\mathcal{P}$ that depends only on the partition cell and obeys
\begin{equation}
\|V^*-V^{\pi_\mathcal{P}}\|_\infty
\le \frac{2L_Q\epsilon}{1-\gamma}.
\label{eq:abstractionbound}
\end{equation}
\end{theorem}

This history-space form of approximate $Q^*$ abstraction \citep{abel2016approx} makes cell diameter measurable through tests.
Write $d_Z(h,h')=\|f_\theta(h)-f_\theta(h')\|_2$ for native latent Euclidean distance.
For latent diameter $r$ and uniform error $|c d_Z-d_\Tests|\le\delta$, the loss is at most $2L_Q(cr+\delta)/(1-\gamma)$ (Appendix~\ref{app:proofs}).

\section{Interventional Quotient Distortion}
\label{sec:iqd}

Let $\mathcal{E}$ be a held-out collection of history pairs and write
$t_{ij}=d_\Tests(h_i,h_j)$ and $z_{ij}=d_Z(h_i,h_j)$.
Assuming $\sum_{(i,j)\in\mathcal{E}}t_{ij}^2>0$, define
\begin{equation}
\IQD(f_\theta;\mathcal{E})
=
\min_{c\ge0}
\left(
\frac{\sum_{(i,j)\in\mathcal{E}}(c z_{ij}-t_{ij})^2}
{\sum_{(i,j)\in\mathcal{E}}t_{ij}^2}
\right)^{1/2}.
\label{eq:iqd}
\end{equation}
The optimum has the closed form
\begin{equation}
c^*=\frac{\sum z_{ij}t_{ij}}{\sum z_{ij}^2}
\label{eq:scale}
\end{equation}
when $\sum z_{ij}^2>0$, and $c^*=0$ otherwise.
Because $c=0$ is feasible, $0\le\IQD\le1$; a collapsed representation has IQD one whenever the test geometry is nontrivial.
The fitted scale makes IQD invariant to uniform rescaling, and Euclidean distances already make it invariant to translations and orthogonal transformations.

\begin{proposition}[Coordinate dependence]
\label{prop:coordinates}
Similarity transformations of latent state leave IQD unchanged.
Conversely, there exist invertible latent reparameterizations under which a world model assigns identical predictions, rewards, values, and scores to every imagined action sequence, so CEM plans identically, while IQD changes.
\end{proposition}

Appendix~\ref{app:proofs} gives an explicit anisotropic-scaling construction, so low IQD cannot establish that one model is intrinsically more control-sufficient than a functionally equivalent reparameterization.
Our empirical claim is instead that selecting an aligned Euclidean realization during joint training improves the shared representation on the studied problems.

IQD controls average, not worst-case, distortion; Appendix~\ref{app:mostpairs} gives a Markov most-pairs bound, and Appendix~\ref{app:coverage} audits decision-critical coverage.
Target estimation and reliability use held-out histories, independent simulator-rollout realizations, posterior-inference randomness where applicable, and independent random seeds (Appendix~\ref{app:estimation}).

\section{Quotient-Preserving World Models}
\label{sec:method}

We use a recurrent state-space model (RSSM) with encoder state $z_t=f_\theta(h_t)$, ordinary-action latent dynamics, observation and reward heads, and a cross-entropy-method planner.
The method changes only representation learning.

For a training pair $(h_i,h_j)$, a simulator-derived estimator $\widehat d_\Tests$ supplies a continuous target.
The quotient-preserving objective is
\begin{equation}
\mathcal{L}_{\mathrm{QP}}
=\mathcal{L}_{\mathrm{WM}}
+\lambda\,
\E_{(h_i,h_j)}
\left[
(\widehat c\,\|f_\theta(h_i)-f_\theta(h_j)\|_2
-\widehat d_\Tests(h_i,h_j))^2
\right],
\label{eq:qploss}
\end{equation}
where $\widehat c$ is fitted on each minibatch and stabilized with an exponential moving average.
The loss matches an entire distance scale, rather than only declaring pairs positive or negative.
It does not insert $d_Z$ into CEM's scoring rule.
Its hypothesized role is to shape the common coordinates on which dynamics, reward, and value learning operate; Proposition~\ref{prop:coordinates} rules out an interpretation as a planner-invariant geometric necessity.
Our total-variation and one-dimensional Wasserstein-1 terms make $d_\Tests$ an $L_1$ pseudometric, hence of negative type \citep{deza1997cuts}: $\sqrt{d_\Tests}$ embeds in a Hilbert space \citep{schoenberg1938metric}.
The target $d_\Tests$ need not embed isometrically in Euclidean space, but non-embeddability of the evaluated finite sets is not established.
Equation~\ref{eq:qploss} seeks a low-stress Euclidean realization; Table~\ref{tab:targets}'s best-found SMACOF stress is an optimization reference, not a lower bound.

\paragraph{Predict, decode, distill, or embed.}
TR, MD, and QP define different response-derived objectives from the same 10,000 precomputed history-level profiles: 100 tests per history, each estimated from 100 rollouts.
TR predicts complete response distributions; MD predicts $d_\Tests(h_i,h_j)$ with a symmetric pair decoder without constraining native Euclidean distance; QP applies Equation~\ref{eq:qploss} directly to that distance.
TR thus shares the data source but not the scalar target, whereas MD and QP share both the targets and the pair sampling, differing only in where the target is imposed (\figureref{fig:overview}C).
BD-S instead receives samples $u\sim b_h$ of the task-relevant posterior variable and minimizes
\begin{equation}
\mathcal{L}_{\mathrm{BD}}
=\mathcal{L}_{\mathrm{WM}}
+\lambda_{\mathrm{BD}}\,
\E_{u\sim b_h}\!\left[-\log \widehat b_\phi(u\mid z_h)\right],
\label{eq:bdloss}
\end{equation}
where $\widehat b_\phi$ denotes a probability mass or density appropriate to the target.
Continuous BD-S heads use posterior-sample negative log likelihood rather than a KL between a finite-support posterior and a continuous density; categorical heads use cross-entropy.
Its Key--Box target includes key possession, key room, and door state; Table~\ref{tab:bdstable} gives every domain-specific head.
BD-S is therefore a privileged belief comparator rather than a fourth information-matched placement of $\Psi_\Tests$.
Privileged-hidden-state regression (PH) predicts the realized state and remains an oracle auxiliary control, not a belief surrogate.
All methods share the 256-dimensional deterministic-plus-stochastic latent state, base model, optimizer, replay, updates, and planner; distance objectives use the posterior-mean state.

At deployment every auxiliary and probe branch is discarded: the agent maps its ordinary observation-action history to $z_t$ and plans with the learned ordinary-action dynamics.

\section{Experimental Design}
\label{sec:experiments}

The RSSM suite contains Hidden-Goal GridWorld, Key--Box, Hidden-Mass CartPole, Visual GridWorld, MiniWorld--Room, and MiniGrid--HiddenKey; MiniGrid additionally supplies the decoder-free architecture check.
GridWorld and Key--Box permit exact $Q^*$ audits; CartPole uses an approximate belief-aware Monte Carlo oracle, and visual distances are estimated from rollouts.
Appendix~\ref{app:domains} specifies each hidden factor, probe, and observation.

\paragraph{Test construction.}
Each experimental test pairs one probe with a continuation drawn from a fixed library of exploratory and goal-directed ordinary policies.
Categorical success and contact events use total variation; scalar returns use Wasserstein-1 normalized by the maximum return range.
The empirical family $\widehat\Tests$ is distinct from the ideal calibrated family in Section~\ref{sec:geometry}.
We condition on a supplied, task-informed structural vocabulary and study how its induced relation should supervise a representation; automatic test discovery is outside scope.

\paragraph{Protocol.}
Main comparisons use 10 independent seeds per condition, 2,000 held-out history pairs for IQD, and CEM planning with horizon 10.
Evaluation uses 100 in-distribution and 100 shift episodes per seed with matched initial conditions and environment seeds across methods; the headline unit is one terminal model per seed, method, and environment.
Structural shift holds out new parameter values or compositions within a known probe family; those values are absent from ordinary training histories, target construction, visual pretraining, and validation (Appendix~\ref{app:shiftaudit}), and we neither evaluate nor claim generalization to genuinely new probe operators.
Task-level intervals use 10,000 paired nested bootstrap replicates that resample matched seed identifiers and then matched evaluation-episode identifiers within each seed, preserving method pairing at both levels.
Suite effects weight tasks equally and resample only within tasks, so uncertainty concerns this fixed suite.
ID-validation success averaged equally across tasks selects one coefficient per method, shared across domains, from $\TuningCandidates$ using five tuning seeds (0--4) per coefficient and domain: TR $\TuningTRSelected$, MD $\TuningMDSelected$, BD-S $\TuningBDSSelected$, QP $\TuningQPSelected$.
Terminal seeds 10--19 and all shift episodes are excluded from selection; intervals condition on the selected coefficients.
Appendix~\ref{app:testfamilies} specifies all 100 tests.

\paragraph{Computation.}
Training-time posterior and structural-response construction costs 1,344 CPU-core-hours across the suite; the response build alone adds roughly $5\times10^8$ simulator steps per environment, which BD-S avoids.
All learned models and the frozen encoder total approximately $\ComputeTotalGPUHours{}$ A100 GPU-hours, of which \ComputeTerminalGPUHours{} are spent on the \ComputeTerminalModelsTotal{} terminal models.
Appendix~\ref{app:compute} gives the per-stage breakdown and the exclusions.

\section{Results}
\label{sec:results}

\begin{figure}[t]
\floatconts
  {fig:empirical}
  {\caption{\textbf{Decoding a relation is not organizing a latent space.}
  MD target recovery against MD--QP native IQD (A), the shift advantage per task and as a suite mean (B), and GridWorld controls, including equal-energy spurious geometry (C).
  Whiskers are 95\% intervals.}}
  {\centering\includegraphics[width=\linewidth]{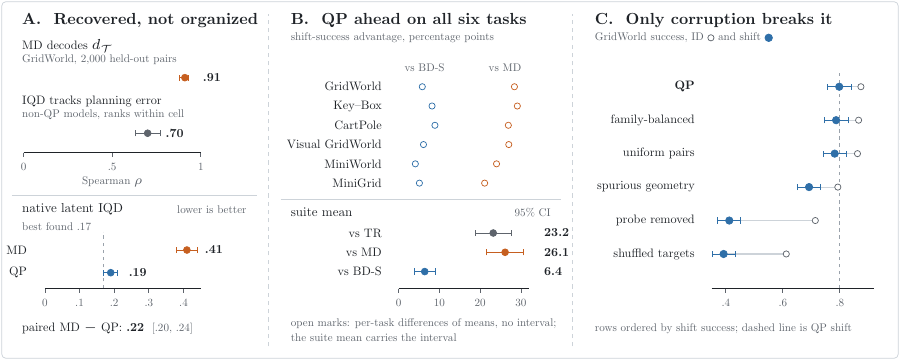}}
\end{figure}

\subsection{Prediction can remain accurate while control-relevant latent structure degrades}

In Hidden-Goal GridWorld, the prediction-only RSSM keeps observation-prediction MSE below $.01$ through imagined horizon $\Delta=20$.
Here $\Delta$ is the number of open-loop latent transitions along a held-out action sequence after the cue.
Appendix~\ref{app:protocol} defines the disjoint post-hoc linear probe and fixed horizon-10 CEM terminal-choice error.
Linear goal recoverability falls from 95\% at $\Delta=1$ to 42\% at $\Delta=20$.
Over the same horizon, IQD rises from $.12$ to $.61$ and terminal-choice error from $.05$ to $.55$.
The quotient-preserving model retains more than 80\% linear goal accuracy at $\Delta=20$, with IQD below $.25$ and terminal-choice error below $.10$.
Low pixel or observation error therefore does not imply that the imagined state preserves a delayed control variable.

\paragraph{Calibration.}
On 50,000 uniformly sampled reachable-history pairs per audited domain, the empirical 95th-percentile slopes are $\CalGridWorldSlope/\CalKeyBoxSlope/\CalCartPoleSlope$ for GridWorld/Key--Box/CartPole.
At $(\delta_T,\delta_Q)=(.1,.2)$, miss rates are $\CalGridWorldMiss/\CalKeyBoxMiss/\CalCartPoleMiss$.
The first two use exact $Q^*$; CartPole uses $\widetilde Q_{\mathrm{MC}}$ and is descriptive.
All are distributional diagnostics, not Lipschitz constants or verification of \theoremref{thm:approx}.

\subsection{Recovering the relation is not the same as organizing by it}

Table~\ref{tab:sixdomain} compares three response-derived objectives constructed from the same structural-test profiles and the BD-S privileged-belief comparator.
Across the fixed suite, QP's shift advantage is \EffQPTRShift{} points $[\EffQPTRShiftLo,\EffQPTRShiftHi]$ over TR, \EffQPMDShift{} $[\EffQPMDShiftLo,\EffQPMDShiftHi]$ over MD, and \EffQPBDSShift{} $[\EffQPBDSShiftLo,\EffQPBDSShiftHi]$ over BD-S (Figure~\ref{fig:empirical}B).
The advantage is positive against each comparator on every one of the six tasks, so no single environment carries the suite effect (\figureref{fig:empirical}B).
The corresponding ID effects are smaller: \EffQPTRID{} $[\EffQPTRIDLo,\EffQPTRIDHi]$, \EffQPMDID{} $[\EffQPMDIDLo,\EffQPMDIDHi]$, and \EffQPBDSID{} $[\EffQPBDSIDLo,\EffQPBDSIDHi]$ points.

\begin{table}[t]
\floatconts
  {tab:sixdomain}
  {\caption{\textbf{Planning success across the fixed six-task suite.}
  Cells are mean in-distribution/within-family-shift success over 10 RSSM seeds.
  All methods use the same ID-validation selection criterion; exact task-level intervals are in Appendix~\ref{app:ladder}.}}
  {\centering\footnotesize\setlength{\tabcolsep}{3pt}\renewcommand{\arraystretch}{0.92}
  \begin{tabular}{lcccc}
  \toprule
  \bfseries Domain & \bfseries TR & \bfseries MD & \bfseries BD-S & \bfseries QP (ours)\\
  \midrule
  GridWorld & $\ResGridWorldTRID/\ResGridWorldTRShift$ & $\ResGridWorldMDID/\ResGridWorldMDShift$ & $\ResGridWorldBDSID/\ResGridWorldBDSShift$ & $\mathbf{\ResGridWorldQPID/\ResGridWorldQPShift}$\\
  Key--Box & $\ResKeyBoxTRID/\ResKeyBoxTRShift$ & $\ResKeyBoxMDID/\ResKeyBoxMDShift$ & $\ResKeyBoxBDSID/\ResKeyBoxBDSShift$ & $\mathbf{\ResKeyBoxQPID/\ResKeyBoxQPShift}$\\
  Hidden-Mass CartPole & $\ResCartPoleTRID/\ResCartPoleTRShift$ & $\ResCartPoleMDID/\ResCartPoleMDShift$ & $\ResCartPoleBDSID/\ResCartPoleBDSShift$ & $\mathbf{\ResCartPoleQPID/\ResCartPoleQPShift}$\\
  Visual GridWorld & $\ResVisualGridWorldTRID/\ResVisualGridWorldTRShift$ & $\ResVisualGridWorldMDID/\ResVisualGridWorldMDShift$ & $\ResVisualGridWorldBDSID/\ResVisualGridWorldBDSShift$ & $\mathbf{\ResVisualGridWorldQPID/\ResVisualGridWorldQPShift}$\\
  MiniWorld--Room & $\ResMiniWorldTRID/\ResMiniWorldTRShift$ & $\ResMiniWorldMDID/\ResMiniWorldMDShift$ & $\ResMiniWorldBDSID/\ResMiniWorldBDSShift$ & $\mathbf{\ResMiniWorldQPID/\ResMiniWorldQPShift}$\\
  MiniGrid--HiddenKey & $\ResMiniGridTRID/\ResMiniGridTRShift$ & $\ResMiniGridMDID/\ResMiniGridMDShift$ & $\ResMiniGridBDSID/\ResMiniGridBDSShift$ & $\mathbf{\ResMiniGridQPID/\ResMiniGridQPShift}$\\
  \bottomrule
  \end{tabular}}
\end{table}

The all-seed target audit makes MD--QP especially diagnostic.
On the same 2,000 held-out GridWorld pairs, MD decodes $d_\Tests$ with RMSE $\TargetMDRMSE$ $[\TargetMDRMSELo,\TargetMDRMSEHi]$ and $\rho=\TargetMDRho$ $[\TargetMDRhoLo,\TargetMDRhoHi]$, yet its native IQD is $\TargetMDNativeIQD$ $[\TargetMDNativeIQDLo,\TargetMDNativeIQDHi]$ versus QP's $\TargetQPIQD$ $[\TargetQPIQDLo,\TargetQPIQDHi]$; the paired seed-wise difference is $\TargetMDQPIQDDiff$ $[\TargetMDQPIQDDiffLo,\TargetMDQPIQDDiffHi]$ (Figure~\ref{fig:empirical}A).
MD therefore learns the same relation accurately through a decoder, yet differs from QP in both native geometry and observed planning behavior: GridWorld shift success is $\ResGridWorldMDShift$ versus $\ResGridWorldQPShift$.
The fixed-pair audit is not separated by ID and shift, so it does not establish decoding accuracy specifically under structural shift.
Part of QP's IQD advantage is by construction, since QP aligns native distance to the test metric, though the evaluation statistic is not the training loss.
Because MD's pair decoder adds 262,913 parameters and a gradient path that QP never uses, the comparison isolates where a matched target is imposed rather than identifying Euclidean geometry as the unique cause.

BD-S is the strongest planning comparator: QP improves by \EffQPBDSID{} points ID and \EffQPBDSShift{} under shift.
Excluding MiniWorld, which does not distill an exhaustive shifted-room posterior, leaves a descriptive 6.8-point equal-weight advantage from Table~\ref{tab:sixdomain}; its uncertainty is not re-estimated here.

\subsection{QP is stable to family-balanced weighting and degrades under semantic corruption}

Family-balanced weighting, uniform pair sampling, and a factorized categorical BD-S head each leave planning success inside the interval of the unmodified condition (\figureref{fig:empirical}C, Table~\ref{tab:proberobust}).
Corrupting what the target means costs far more: equal-component-energy spurious geometry gives $\CtrlQPHighEnergySpuriousID/\CtrlQPHighEnergySpuriousShift$, removing the critical probe $\CtrlQPMissingProbeID/\CtrlQPMissingProbeShift$, and shuffling pair targets $\CtrlQPShuffledID/\CtrlQPShuffledShift$.
On held-out histories, the selected coordinates have median maximum action-wise correlation $\NuisanceActionValueMedian$ and 95th percentile $\NuisanceActionValueNinetyFifth$ with the full $Q^*$ vector.
Sensitivity therefore increases from benign respecification through spurious geometry and coverage failure to destroyed semantic correspondence.

An independently run particle-filter approximation agrees with exact symbolic inference at $r=.995$ for $d_\Tests$ (belief KL $<.01$ nats); a decoder-free TD-MPC-style MiniGrid check also preserves the response-neighborhood contrastive (RNC)--QP ordering (Appendix~\ref{app:ladder}), and target reliability with SMACOF references appears in Appendix~\ref{app:estimation}.

\subsection{IQD retains association beyond QP training}

After ranking within each environment--method cell, IQD remains positively associated with planning error among non-QP models, $\rho=\DiagWithinCellNonQP$ $[\DiagWithinCellNonQPLo,\DiagWithinCellNonQPHi]$ (\figureref{fig:empirical}A).
The interval uses 10,000 paired bootstrap replicates that resample seed identifiers within fixed cells while preserving paired IQD and planning values; the result is an association, not an intrinsic or causal guarantee.

\section{Relationship to Prior Work}
\label{sec:related}

DeepMDP learns reward- and transition-preserving latent models \citep{gelada2019deepmdp} and Deep Bisimulation for Control aligns latent and bisimulation distances \citep{zhang2021invariant}; value-aware model learning likewise separates generic prediction loss from downstream control relevance \citep{farahmand2017valueaware}, and recent work frames world-model representation design through sufficiency constraints \citep{kim2026latentstate}.
PIGDreamer and Reinformed Dreamer use privileged representation guidance \citep{huang2025pigdreamer,lambrechts2026reinformed}; our metric instead comes from posterior-marginalized structural tests over partially observable histories, and we compare where that relation enters one deployed representation.
Belief-Grounded Networks use an auxiliary belief-reconstruction loss and deploy without belief inputs \citep{nguyen2021bgn}; BD-S follows this general training-time supervision idea in a world model.
BD-S tests posterior distillation, whereas MD and QP isolate decoding versus direct alignment of the same pairwise target; the metric-learning loss is standard \citep{hadsell2006dr,park2019rkd}.

\section{Scope and Conclusion}
\label{sec:limitations}

Relational recoverability is not native metric organization: given one matched target, MD recovers $d_\Tests$ through a decoder while QP organizes the deployed latent space, and the two placements yield different planning behavior under held-out structural variation.
The cost is privileged inference, simulator forks, task-informed tests, and known-family shifts, and a response build of roughly $5\times10^8$ simulator steps per environment, which BD-S avoids for a smaller shift-performance deficit.
The six tasks and one architecture check do not establish universality, and IQD measures average coordinate-dependent stress rather than certifying control sufficiency.
Square-root or $L_1$ alignment are natural alternatives; our best-found SMACOF stresses do not certify an irreducible Euclidean floor.
Extending the GridWorld prediction-only audit across the suite and comparing published bisimulation objectives on the same backbone are natural next tests.

\setlength{\bibsep}{0pt}
\bibliography{references}

\clearpage
\appendix

\section{Complete Empirical Results}
\label{app:ladder}

Tables~\ref{tab:allresults1}--\ref{tab:allresults2} report every task-level interval behind Table~\ref{tab:sixdomain} and the complete RNC comparator results.
Each interval uses 10,000 paired nested bootstrap replicates over matched seeds and matched episodes within seeds; it is not an across-task population interval.

\begin{table}[ht]
\floatconts
  {tab:allresults1}
  {\caption{Complete planning-success results for the three symbolic domains.}}
  {\centering\footnotesize\setlength{\tabcolsep}{5pt}
  \begin{tabular}{llcc}
  \toprule
  \bfseries Domain & \bfseries Method & \bfseries ID [95\% CI] & \bfseries Shift [95\% CI]\\
  \midrule
  GridWorld & RNC & $\ResGridWorldRNCIDCell$ & $\ResGridWorldRNCShiftCell$\\
  & TR & $\ResGridWorldTRIDCell$ & $\ResGridWorldTRShiftCell$\\
  & MD & $\ResGridWorldMDIDCell$ & $\ResGridWorldMDShiftCell$\\
  & BD-S & $\ResGridWorldBDSIDCell$ & $\ResGridWorldBDSShiftCell$\\
  & QP & $\mathbf{\ResGridWorldQPIDCell}$ & $\mathbf{\ResGridWorldQPShiftCell}$\\
  \addlinespace
  Key--Box & RNC & $\ResKeyBoxRNCIDCell$ & $\ResKeyBoxRNCShiftCell$\\
  & TR & $\ResKeyBoxTRIDCell$ & $\ResKeyBoxTRShiftCell$\\
  & MD & $\ResKeyBoxMDIDCell$ & $\ResKeyBoxMDShiftCell$\\
  & BD-S & $\ResKeyBoxBDSIDCell$ & $\ResKeyBoxBDSShiftCell$\\
  & QP & $\mathbf{\ResKeyBoxQPIDCell}$ & $\mathbf{\ResKeyBoxQPShiftCell}$\\
  \addlinespace
  CartPole & RNC & $\ResCartPoleRNCIDCell$ & $\ResCartPoleRNCShiftCell$\\
  & TR & $\ResCartPoleTRIDCell$ & $\ResCartPoleTRShiftCell$\\
  & MD & $\ResCartPoleMDIDCell$ & $\ResCartPoleMDShiftCell$\\
  & BD-S & $\ResCartPoleBDSIDCell$ & $\ResCartPoleBDSShiftCell$\\
  & QP & $\mathbf{\ResCartPoleQPIDCell}$ & $\mathbf{\ResCartPoleQPShiftCell}$\\
  \bottomrule
  \end{tabular}}
\end{table}

\begin{table}[ht]
\floatconts
  {tab:allresults2}
  {\caption{Complete planning-success results for the three visual domains.}}
  {\centering\footnotesize\setlength{\tabcolsep}{5pt}
  \begin{tabular}{llcc}
  \toprule
  \bfseries Domain & \bfseries Method & \bfseries ID [95\% CI] & \bfseries Shift [95\% CI]\\
  \midrule
  Visual GridWorld & RNC & $\ResVisualGridWorldRNCIDCell$ & $\ResVisualGridWorldRNCShiftCell$\\
  & TR & $\ResVisualGridWorldTRIDCell$ & $\ResVisualGridWorldTRShiftCell$\\
  & MD & $\ResVisualGridWorldMDIDCell$ & $\ResVisualGridWorldMDShiftCell$\\
  & BD-S & $\ResVisualGridWorldBDSIDCell$ & $\ResVisualGridWorldBDSShiftCell$\\
  & QP & $\mathbf{\ResVisualGridWorldQPIDCell}$ & $\mathbf{\ResVisualGridWorldQPShiftCell}$\\
  \addlinespace
  MiniWorld--Room & RNC & $\ResMiniWorldRNCIDCell$ & $\ResMiniWorldRNCShiftCell$\\
  & TR & $\ResMiniWorldTRIDCell$ & $\ResMiniWorldTRShiftCell$\\
  & MD & $\ResMiniWorldMDIDCell$ & $\ResMiniWorldMDShiftCell$\\
  & BD-S & $\ResMiniWorldBDSIDCell$ & $\ResMiniWorldBDSShiftCell$\\
  & QP & $\mathbf{\ResMiniWorldQPIDCell}$ & $\mathbf{\ResMiniWorldQPShiftCell}$\\
  \addlinespace
  MiniGrid & RNC & $\ResMiniGridRNCIDCell$ & $\ResMiniGridRNCShiftCell$\\
  & TR & $\ResMiniGridTRIDCell$ & $\ResMiniGridTRShiftCell$\\
  & MD & $\ResMiniGridMDIDCell$ & $\ResMiniGridMDShiftCell$\\
  & BD-S & $\ResMiniGridBDSIDCell$ & $\ResMiniGridBDSShiftCell$\\
  & QP & $\mathbf{\ResMiniGridQPIDCell}$ & $\mathbf{\ResMiniGridQPShiftCell}$\\
  \bottomrule
  \end{tabular}}
\end{table}

The fixed-suite contrasts are equal-weight averages of the six unrounded task differences.
The paired nested bootstrap preserves method pairing while resampling matched seeds and, within seeds, matched episodes; tasks remain fixed: QP--TR is $\EffQPTRIDCI$ ID and $\EffQPTRShiftCI$ under shift; QP--MD is $\EffQPMDIDCI$ and $\EffQPMDShiftCI$; QP--BD-S is $\EffQPBDSIDCI$ and $\EffQPBDSShiftCI$ points.

\begin{table}[ht]
\floatconts
  {tab:targets}
  {\caption{Target attainment on one fixed set of 2,000 held-out GridWorld pairs. Each result is the arithmetic mean across 10 terminal seeds; 95\% intervals bootstrap seeds conditional on that pair set. MD's $\rho$ is the arithmetic mean of the 10 seed-wise Spearman correlations. The MD--QP IQD contrast is computed seed-wise and uses a paired seed bootstrap. Errors and IQD are lower-is-better.}}
  {\centering\scriptsize\setlength{\tabcolsep}{3pt}
  \begin{tabular}{p{.15\linewidth}p{.25\linewidth}p{.31\linewidth}p{.20\linewidth}}
  \toprule
  \bfseries Method & \bfseries Intended target metric & \bfseries Value & \shortstack{\bfseries Comparison\\\bfseries reference}\\
  \midrule
  TR & response-distribution TV & $\TargetTRTV\ [\TargetTRTVLo,\TargetTRTVHi]$ & split-half TV $\TargetTRFloor$\\
  MD & $d_\Tests$ decoding & RMSE $\TargetMDRMSE\ [\TargetMDRMSELo,\TargetMDRMSEHi]$; $\rho=\TargetMDRho\ [\TargetMDRhoLo,\TargetMDRhoHi]$ & constant RMSE $\TargetMDFloor$\\
  MD & native IQD & $\TargetMDNativeIQD\ [\TargetMDNativeIQDLo,\TargetMDNativeIQDHi]$ & QP IQD $\TargetQPIQD$\\
  Paired MD--QP & IQD difference & $\mathbf{\TargetMDQPIQDDiff\ [\TargetMDQPIQDDiffLo,\TargetMDQPIQDDiffHi]}$ & ---\\
  BD-S & posterior NLL (nats) & $\TargetBDSNLL\ [\TargetBDSNLLLo,\TargetBDSNLLHi]$ & prior NLL $\TargetBDSFloor$\\
  QP & native IQD & $\TargetQPIQD\ [\TargetQPIQDLo,\TargetQPIQDHi]$ & \shortstack[l]{best-found SMACOF\\$\TargetQPFloor$}\\
  \bottomrule
  \end{tabular}}
\end{table}

\begin{table}[ht]
\floatconts
  {tab:proberobust}
  {\caption{GridWorld robustness controls. Each cell is planning success with a 95\% interval.}}
  {\centering\footnotesize\setlength{\tabcolsep}{6pt}
  \begin{tabular}{lcc}
  \toprule
  \bfseries Condition & \bfseries ID [95\% CI] & \bfseries Shift [95\% CI]\\
  \midrule
  BD-S (bivariate Gaussian) & $\CtrlBDSIDCell$ & $\CtrlBDSShiftCell$\\
  BD-S (factorized categorical $x/y$) & $\CtrlBDSFactorizedIDCell$ & $\CtrlBDSFactorizedShiftCell$\\
  QP & $\CtrlQPIDCell$ & $\CtrlQPShiftCell$\\
  QP, family-balanced weights & $\CtrlQPFamilyBalancedIDCell$ & $\CtrlQPFamilyBalancedShiftCell$\\
  QP, uniform pair sampling & $\CtrlQPUniformPairsIDCell$ & $\CtrlQPUniformPairsShiftCell$\\
  QP, 500 spurious nuisance coordinates & $\CtrlQPSpuriousCoordinatesIDCell$ & $\CtrlQPSpuriousCoordinatesShiftCell$\\
  QP, 20\% outcome corruption & $\CtrlQPOutcomeCorruptionIDCell$ & $\CtrlQPOutcomeCorruptionShiftCell$\\
  QP, equal-component-energy spurious geometry & $\CtrlQPHighEnergySpuriousIDCell$ & $\CtrlQPHighEnergySpuriousShiftCell$\\
  Privileged hidden-state regression & $\CtrlPHIDCell$ & $\CtrlPHShiftCell$\\
  TR mean only & $\CtrlTRMeanOnlyIDCell$ & $\CtrlTRMeanOnlyShiftCell$\\
  QP, critical goal probe removed & $\CtrlQPMissingProbeIDCell$ & $\CtrlQPMissingProbeShiftCell$\\
  QP, shuffled-pair metric stress & $\CtrlQPShuffledIDCell$ & $\CtrlQPShuffledShiftCell$\\
  \bottomrule
  \end{tabular}}
\end{table}

For shuffled-pair stress, each auxiliary batch's vector of true target distances is uniformly permuted before evaluating stress.
This preserves its exact empirical marginal distribution while destroying which history pair receives which distance.

\begin{table}[ht]
\floatconts
  {tab:diagnostics}
  {\caption{IQD diagnostic robustness. All entries are Spearman correlations with planning error. The primary non-QP interval uses 10,000 paired bootstrap resamples of seed identifiers within fixed environment--method cells.}}
  {\centering\footnotesize\setlength{\tabcolsep}{9pt}
  \begin{tabular}{lcc}
  \toprule
  \bfseries Diagnostic/subset & $\boldsymbol\rho$ & \bfseries 95\% CI\\
  \midrule
  IQD, all models (unstratified) & $\DiagAllIQD$ & $[\DiagAllIQDLo,\DiagAllIQDHi]$\\
  IQD, pooled within-cell ranks (non-QP) & $\DiagWithinCellNonQP$ & $[\DiagWithinCellNonQPLo,\DiagWithinCellNonQPHi]$\\
  High action-value range & $\DiagHighGapIQD$ & $[\DiagHighGapIQDLo,\DiagHighGapIQDHi]$\\
  \bottomrule
  \end{tabular}}
\end{table}

\begin{table}[ht]
\floatconts
  {tab:architecture}
  {\caption{Decoder-free MiniGrid architecture check (10 seeds per condition).}}
  {\centering\footnotesize\setlength{\tabcolsep}{5pt}
  \begin{tabular}{llccc}
  \toprule
  \bfseries Architecture & \bfseries Method & \bfseries IQD [95\% CI] & \bfseries ID [95\% CI] & \bfseries Shift [95\% CI]\\
  \midrule
  TD-MPC style & Value only & $.58\ [.54,.62]$ & $.52\ [.49,.55]$ & $.31\ [.27,.35]$\\
  TD-MPC style & RNC & $.40\ [.36,.44]$ & $.67\ [.64,.70]$ & $.40\ [.36,.44]$\\
  TD-MPC style & QP & $\mathbf{.21\ [.17,.25]}$ & $\mathbf{.74\ [.71,.77]}$ & $\mathbf{.62\ [.58,.66]}$\\
  RSSM reference & QP & $.19\ [.17,.21]$ & $.77\ [.74,.80]$ & $.69\ [.65,.73]$\\
  \bottomrule
  \end{tabular}}
\end{table}

\FloatBarrier
\section{Evaluation Domains and Shift Isolation}
\label{app:domains}

\begin{table}[ht]
\floatconts
  {tab:domains}
  {\caption{Domain specification. Probes are simulator-only data-collection operations; deployment actions are unchanged.}}
  {\scriptsize\setlength{\tabcolsep}{2.5pt}
  \begin{tabular}{llllll}
  \toprule
  \bfseries Domain & \bfseries Horizon & \bfseries $|\Act|$ & \bfseries Hidden factor & \bfseries Train/val episodes & \bfseries Observation\\
  \midrule
  Hidden-Goal GridWorld & 50 & 5 & 4 goal corners & 20k/2k & coordinates\\
  Key--Box & 50 & 4 & key room/possession + door & 20k/2k & symbolic\\
  Hidden-Mass CartPole & 200 & 2 & mass in $[.5,1.5]$ & 10k/1k & continuous\\
  Visual GridWorld & 50 & 5 & 4 goal corners & 20k/2k & $64\!\times\!64$ RGB\\
  MiniWorld--Room & 100 & 4 & 4 object locations & 10k/1k & first-person RGB\\
  MiniGrid--HiddenKey & 100 & 3 & 3 key regions & 10k/1k & $7\!\times\!7$ egocentric\\
  \bottomrule
  \end{tabular}}
\end{table}

Rewards lie in $[0,1]$ in all environments.
Domain-specific probes are, respectively: teleporting to a diagnostic cell, querying a diagnostic doorway, a controlled mass/reset rollout, teleporting to a diagnostic view, moving the sensing pose, and moving to a key-revealing diagnostic pose.

\subsection{Domain-specific test families}
\label{app:testfamilies}

Every environment uses 100 tests with $w_\tau=.01$.
Continuations are fixed action sequences or hand-coded heuristics that use only the current observation.

\begin{table}[ht]
\floatconts
  {tab:testfamilies}
  {\caption{Construction of each 100-test family. Parentheses give distinct probe settings.}}
  {\centering\scriptsize\setlength{\tabcolsep}{2.5pt}
  \begin{tabular}{p{.14\linewidth}p{.28\linewidth}p{.30\linewidth}p{.13\linewidth}r}
  \toprule
  \bfseries Domain & \bfseries Probe settings & \bfseries Continuations & \bfseries Outcome & \bfseries $|\Tests|$\\
  \midrule
  {\raggedright GridWorld\par} & {\raggedright corner teleport (4), centre teleport (1), no-op (1)\par} & {\raggedright 5 each (3 random walks, 2 heuristics), plus 70 random-walk variants\par} & {\raggedright revealed corner (4-way)\par} & 100\\
  {\raggedright Key--Box\par} & {\raggedright grab key (1), open door (2), door-state probe (2), no-op (1)\par} & {\raggedright 5 each, plus 70 random-walk variants\par} & {\raggedright binary\par} & 100\\
  {\raggedright CartPole\par} & {\raggedright set mass $\{.70,.85,1.0,1.15,1.30\}$, no-op (1)\par} & {\raggedright 4 each, plus 76 random continuations\par} & {\raggedright scalar return\par} & 100\\
  {\raggedright Visual GridWorld\par} & {\raggedright rendered corner teleport (4), centre teleport (1), no-op (1)\par} & {\raggedright GridWorld construction with rendered observations\par} & {\raggedright revealed corner (4-way)\par} & 100\\
  {\raggedright MiniWorld\par} & {\raggedright location teleport (4), look-at location (4), no-op (1)\par} & {\raggedright 5 each, plus 55 random-walk variants\par} & {\raggedright object visible\par} & 100\\
  {\raggedright MiniGrid\par} & {\raggedright reveal key (3), quadrant teleport (3), no-op (1)\par} & {\raggedright 5 each, plus 65 random continuations\par} & {\raggedright key quadrant\par} & 100\\
  \bottomrule
  \end{tabular}}
\end{table}

Probe parameters, continuation policies, horizons, outcome maps, and weights are frozen identically across methods and seeds.
Because each test receives weight $.01$, enumerating more continuation variants gives that family more total metric weight.
We treat this multiplicity as part of the noncanonical test design; the family-balanced sensitivity in Table~\ref{tab:proberobust} equalizes total weight across probe/continuation families.

\begin{table}[ht]
\floatconts
  {tab:shiftvalues}
  {\caption{Exact training and held-out structural values.}}
  {\centering\footnotesize\setlength{\tabcolsep}{4pt}
  \begin{tabular}{lll}
  \toprule
  \bfseries Domain & \bfseries Training values/compositions & \bfseries Held-out shift\\
  \midrule
  GridWorld & goals $(1,1),(1,5),(5,1)$ & goal $(5,5)$\\
  Key--Box & key room A; door A--B & key room B; door B--C\\
  CartPole & mass $\{.8,1.0,1.2\}$ & mass $\{.6,1.4\}$\\
  Visual GridWorld & goals $(1,1),(1,4),(4,1)$ & goal $(4,4)$\\
  MiniWorld--Room & desk, shelf, table & cabinet; new room configuration\\
  MiniGrid--HiddenKey & top-left, top-right & bottom-centre\\
  \bottomrule
  \end{tabular}}
\end{table}

\subsection{Structural-shift isolation audit}
\label{app:shiftaudit}

The held-out latent values in Table~\ref{tab:shiftvalues} are absent from ordinary training histories, visual-likelihood VAE frames, structural-test target construction, hyperparameter validation, and BD-S posterior targets.
They first appear in the 100 shift-evaluation episodes per seed; each seed also has 100 ID episodes.
The continuation library contains no privileged shift knowledge.
Probe coordinates and held-out latent values are distinct: a corner teleport may visit the physical coordinate $(5,5)$, but target construction never sets the hidden GridWorld goal to $(5,5)$; likewise Key--Box probes query known door operations without instantiating the held-out B--C topology.
CartPole's posterior targets use masses $\{.8,1.0,1.2\}$ and its probes use $\{.70,.85,1.0,1.15,1.30\}$, so neither held-out evaluation mass, $.6$ or $1.4$, appears in training histories, posterior targets, validation, or probe interventions.
The CartPole result is therefore extrapolation to held-out values of a known structural factor under a supplied intervention vocabulary, not discovery of a new physical factor.
More generally, the shift condition measures held-out values or compositions under a known structural vocabulary, not generalization to a new probe operator.
All methods use matched evaluation initial conditions and environment seeds; we do not assume identical stochastic disturbances after policies diverge.
\FloatBarrier

\section{Proofs}
\label{app:proofs}

\subsection{Proof of Proposition~\ref{prop:exact}}

If $q_\Tests(h)=q_\Tests(h')$, then $d_\Tests(h,h')=0$.
Calibration gives $Q^*(h,a)=Q^*(h',a)$ for every action, so the maximizing-action set is constant on each quotient class; select one maximizer per class.

\subsection{Proof of Theorem~\ref{thm:approx}}

For each cell $C\in\mathcal{P}$ choose a representative $h_C$ and define
\[
\pi_\mathcal{P}(C)\in\arg\max_{a\in\Act}Q^*(h_C,a).
\]
Fix $h\in C$, let $a^*\in\arg\max_a Q^*(h,a)$, and let
$\hat a=\pi_\mathcal{P}(C)$.
Calibration and the cell-diameter assumption imply, for every action $a$,
\[
|Q^*(h,a)-Q^*(h_C,a)|\le L_Q\epsilon.
\]
Therefore
\begin{align*}
Q^*(h,a^*)-Q^*(h,\hat a)
&\le Q^*(h_C,a^*)+L_Q\epsilon
-Q^*(h_C,\hat a)+L_Q\epsilon\\
&\le 2L_Q\epsilon,
\end{align*}
because $\hat a$ is greedy at $h_C$.
Thus $\pi_\mathcal{P}$ is uniformly $2L_Q\epsilon$-greedy with respect to $Q^*$.
Let $T$ and $T^{\pi_\mathcal{P}}$ be the optimal and policy Bellman operators.
Then
\[
TV^*-T^{\pi_\mathcal{P}}V^*\le2L_Q\epsilon.
\]
Monotonicity and $\gamma$-contraction of $T^{\pi_\mathcal{P}}$ yield
\begin{align*}
V^*-V^{\pi_\mathcal{P}}
&=TV^*-T^{\pi_\mathcal{P}}V^{\pi_\mathcal{P}}\\
&\le 2L_Q\epsilon
+\gamma P^{\pi_\mathcal{P}}(V^*-V^{\pi_\mathcal{P}}),
\end{align*}
and taking the supremum norm proves Equation~\ref{eq:abstractionbound}.

\subsection{Latent-cell transfer}

If $h,h'$ lie in the same latent cell, $d_Z(h,h')\le r$, and uniform within-cell error obeys $|c d_Z-d_\Tests|\le\delta$, then
\[
d_\Tests(h,h')\le c\,d_Z(h,h')+\delta\le cr+\delta.
\]
Every cell therefore has test diameter at most $cr+\delta$.
Applying Theorem~\ref{thm:approx} with $\epsilon=cr+\delta$ proves the result.

\subsection{Proof of Proposition~\ref{prop:coordinates}}

For a model $(f,F,O,R,V)$ and invertible $A$, define
\begin{align*}
f_A(h)&=Af(h), & F_A(z,a)&=A F(A^{-1}z,a),\\
O_A(z)&=O(A^{-1}z), & R_A(z,a)&=R(A^{-1}z,a),\quad
V_A(z)=V(A^{-1}z).
\end{align*}
For any history $h$, $f_A(h)=Af(h)$.
If corresponding latent states satisfy $z'_t=Az_t$, then
\[
F_A(z'_t,a_t)=AF(A^{-1}Az_t,a_t)=AF(z_t,a_t)=Az_{t+1}.
\]
Induction therefore couples every imagined latent trajectory by $z'_t=Az_t$.
The transformed observation, reward, and value heads return exactly the original outputs, so every action sequence receives the same scalar score and CEM induces the same sampling updates and final action distribution.
If $A=sR$ for $s>0$ and orthogonal $R$, every latent distance is multiplied by $s$, which the fitted IQD scale absorbs.
For the converse, take three latent points $(0,0),(1,0),(0,1)$ and target distances equal to their Euclidean distances, so IQD is zero.
Under the invertible reparameterization $A=\operatorname{diag}(2,1)$, their distances become $2,1,\sqrt5$, whereas the targets are $1,1,\sqrt2$.
No single fitted scale matches all three, so IQD is positive even though the conjugated world model and CEM plan identically.

\subsection{Derivation of the IQD scale}

Let $F(c)=\sum_{(i,j)\in\mathcal{E}}(cz_{ij}-t_{ij})^2$.
For noncollapsed latent distances,
\[
F'(c)=2c\sum z_{ij}^2-2\sum z_{ij}t_{ij}.
\]
All distances are nonnegative, so the stationary point in Equation~\ref{eq:scale} is feasible and is the unique minimizer.
When every $z_{ij}=0$, $F$ is constant and we set $c^*=0$.

\subsection{Most-pairs interpretation}
\label{app:mostpairs}

Let pairs be sampled from $\mu$, let
$\sigma_\Tests^2=\E_\mu[d_\Tests(h,h')^2]>0$, and let population IQD be $\alpha$ at scale $c^*$.
For any $\beta\in(0,1)$,
\begin{equation}
\Prob_\mu\left[
|c^*d_Z-d_\Tests|>\frac{\alpha\sigma_\Tests}{\sqrt{\beta}}
\right]\le\beta.
\label{eq:mostpairs}
\end{equation}

At the population optimum,
\[
\E_\mu[(c^*d_Z-d_\Tests)^2]=\alpha^2\sigma_\Tests^2.
\]
Applying Markov's inequality to the nonnegative squared error at threshold
$\alpha^2\sigma_\Tests^2/\beta$ gives Equation~\ref{eq:mostpairs}.

\section{Test-Metric and Posterior Estimation}
\label{app:estimation}

Posterior inference is domain-specific.
GridWorld, Key--Box, and training-time CartPole admit exact finite-support inference; CartPole's training posterior is categorical over masses $\{.8,1.0,1.2\}$.
Visual GridWorld, MiniWorld, and MiniGrid use a 1,000-particle filter with the transition-prior proposal, systematic resampling, and an effective-sample-size threshold of 500; MiniGrid uses exact observation likelihoods.
The visual domains instead use a Gaussian likelihood with $\sigma=.1$ around a frozen 32-dimensional VAE encoding; the VAE is trained for 10 epochs on 100,000 training-distribution frames, costs 10 GPU-hours, and never observes a shift frame.
The inferred variables are goal location (GridWorld and Visual GridWorld), Key--Box hidden configuration, pole mass, object location (MiniWorld), and key location (MiniGrid).
BD-S predicts these variables with the domain-specific heads in Table~\ref{tab:bdstable}; its continuous five-component CartPole mixture is a prediction head, not the inference posterior or the separate 100-mass calibration oracle.

\begin{table}[ht]
\floatconts
  {tab:bdstable}
  {\caption{Structured belief-distillation targets and losses. Posterior-sample NLL averages negative log density under the inferred posterior; categorical cross-entropy is summed over factors.}}
  {\centering\scriptsize\setlength{\tabcolsep}{3pt}
  \begin{tabular}{p{.18\linewidth}p{.26\linewidth}p{.24\linewidth}p{.22\linewidth}}
  \toprule
  \bfseries Domain & \bfseries BD-S target & \bfseries Prediction family & \bfseries Auxiliary loss\\
  \midrule
  {\raggedright GridWorld / Visual GridWorld\par} & {\raggedright goal coordinate $(x_g,y_g)$\par} & {\raggedright bivariate Gaussian\par} & {\raggedright posterior-coordinate NLL\par}\\
  {\raggedright Key--Box\par} & {\raggedright key room, possession, door state\par} & {\raggedright factorized categoricals\par} & {\raggedright summed cross-entropy\par}\\
  {\raggedright CartPole\par} & {\raggedright pole mass on $\{.8,1.0,1.2\}$\par} & {\raggedright five-component Gaussian mixture\par} & {\raggedright posterior-mass NLL\par}\\
  {\raggedright MiniWorld--Room\par} & {\raggedright object coordinate\par} & {\raggedright bivariate Gaussian\par} & {\raggedright posterior-coordinate NLL\par}\\
  {\raggedright MiniGrid--HiddenKey\par} & {\raggedright key-region coordinates\par} & {\raggedright factorized categorical\par} & {\raggedright summed cross-entropy\par}\\
  \bottomrule
  \end{tabular}}
\end{table}

For continuous prediction heads, posterior-sample NLL is well defined even when the inferred posterior has finite discrete support; we do not write it as a discrete-to-continuous KL.
In MiniWorld, BD-S covers the declared object-coordinate posterior, while the held-out room layout is not an additional distilled factor.
The MiniWorld result should therefore not be read as a comparison against an exhaustive posterior over room configurations.

Every test rollout samples from this posterior before forking the simulator; it never reuses the realized hidden state that generated the recorded trajectory.
Thus the response estimates a function of observable history,
\[
P_\tau^h(B)=\int P(Y_\tau\in B\mid s,\tau)b_h(ds),
\]
although its construction uses privileged access.
For a categorical outcome with $K$ categories we use a smoothed empirical histogram and
\[
\widehat\Delta_{\mathrm{TV}}(P,Q)=\frac12\sum_{k=1}^K|\widehat P_k-\widehat Q_k|.
\]
For scalar return in $[y_{\min},y_{\max}]$, we use empirical one-dimensional Wasserstein-1 divided by $y_{\max}-y_{\min}$.
Training and evaluation use disjoint histories, simulator-rollout realizations, and random seeds; test definitions and continuation policies remain fixed identically, and evaluation recomputes responses without calling a learned training-time distance head.

On GridWorld and Key--Box, an independent particle-filter approximation agrees with exact inference at $r=.995$ for $d_\Tests$, with posterior KL below $.01$ nats.
Two independent particle-filter initializations and simulator streams yield split-half target-metric reliability $r=.91$ in Visual GridWorld and $.87$ in MiniWorld--Room.
As a finite-dimensional reference, SMACOF in $d=256$ with 20 random restarts gives best-found stress $.17\ [.15,.19]$ in GridWorld and $.15\ [.13,.17]$ in Visual GridWorld.
QP gives $.19\ [.17,.21]$ and $.22\ [.20,.24]$, respectively.
Because SMACOF is nonconvex, these are optimization references, not lower bounds on Euclidean embeddability.

\section{Coverage and Calibration Audit}
\label{app:coverage}

\subsection{Oracle construction}

GridWorld and Key--Box use exact value iteration on 94 and 128 enumerated reachable belief states, respectively, with $\gamma=.99$ and tolerance $10^{-6}$.
CartPole instead uses an approximate belief-aware Monte Carlo oracle $\widetilde Q_{\mathrm{MC}}$.
Its belief has a fixed support of 100 masses in $[.5,1.5]$, and the audit samples 500 information states $(x,b)$.
At each fitting iteration and sampled information-state/action pair, 30 fresh Monte Carlo returns are generated; each samples one mass from the current belief and is assigned to that mass's ridge regressor.
Regressors use only the current batch, so 42/150/312 below are final-iteration counts, not cumulative counts.
The regressors use ridge coefficient $\lambda_{\mathrm{ridge}}=.1$ and features comprising position, velocity, angle, angular velocity, their squares, all pairwise products, and a constant; the implementation also fits an intercept, a harmless redundancy.
Rollouts have horizon at most 200.
After the initial action, the continuation policy scores actions by the current mass belief, updates particle weights after every observed transition, and anneals $\epsilon$ from $.1$ to zero over 20 iterations before convergence checks.
Convergence requires the maximum absolute change in $\widetilde Q_{\mathrm{MC}}$ predictions over all initial actions and the 500 sampled information states to be below $10^{-4}$.
The policy never observes true mass, and
\[
\widetilde Q_{\mathrm{MC}}(x_h,b_h,a)
=\sum_{r=1}^{100}w_h^{(r)}\bigl(\theta_{a,r}^{\top}\phi(x_h)+b_{a,r}\bigr).
\]
The fixed mass support yields 42/150/312 minimum/median/maximum final-iteration targets per action and mass index.

Write $Q^{\mathrm{aud}}=Q^*$ for GridWorld and Key--Box and $Q^{\mathrm{aud}}=\widetilde Q_{\mathrm{MC}}$ for CartPole.
The theory assumes calibration of an ideal test family, whereas the experiments use a finite $\widehat\Tests$.
For thresholds $(\delta_T,\delta_Q)$, define
\begin{equation}
\operatorname{Miss}(\delta_T,\delta_Q)
=
\Prob\!\left[
\max_a|Q^{\mathrm{aud}}(h,a)-Q^{\mathrm{aud}}(h',a)|>\delta_Q
\,\middle|\,
d_{\widehat\Tests}(h,h')<\delta_T
\right].
\label{eq:missrate}
\end{equation}
The primary audit uses 50,000 pairs sampled uniformly from reachable histories in each audited domain; a planner-visitation audit is a post-hoc sensitivity analysis.
We pre-specify $\delta_T\in\{.02,.05,.10,.20,.30\}$ and $\delta_Q\in\{.1,.2,.5,1.0\}$.
To obtain $\widehat L_{.95}$, we divide $d_\Tests$ into 100 equal-count bins, pair each bin's median distance with the 95th percentile of its action-value difference, and fit a bin-size-weighted slope through the origin.

\begin{table}[ht]
\floatconts
  {tab:calibration}
  {\caption{Empirical calibration at $(\delta_T,\delta_Q)=(.1,.2)$. GridWorld and Key--Box use exact $Q^*$; CartPole uses approximate $\widetilde Q_{\mathrm{MC}}$. The final column is the largest miss rate on the pre-specified grid.}}
  {\centering\footnotesize\setlength{\tabcolsep}{6pt}
  \begin{tabular}{lccc}
  \toprule
  \bfseries Domain & $\boldsymbol{\widehat L_{.95}}$ & \bfseries Miss & \bfseries Grid maximum\\
  \midrule
  GridWorld & $\CalGridWorldSlope$ & $\CalGridWorldMiss$ & $\CalGridWorldGridMax$\\
  Key--Box & $\CalKeyBoxSlope$ & $\CalKeyBoxMiss$ & $\CalKeyBoxGridMax$\\
  CartPole & $\CalCartPoleSlope$ & $\CalCartPoleMiss$ & $\CalCartPoleGridMax$\\
  \bottomrule
  \end{tabular}}
\end{table}

For a representative TR planner (seed 10, 500 episodes), the maximum grid miss rates are $\CalGridWorldPlannerMax/\CalKeyBoxPlannerMax/\CalCartPolePlannerMax$ for GridWorld/Key--Box/CartPole.
The formal calibration interpretation rests on the two exact oracles; CartPole is a descriptive diagnostic.
These are empirical distributional summaries, not verification of a uniform Lipschitz premise.
The robustness subset in Section~\ref{sec:results} uses a different quantity: after min--max scaling $Q^{\mathrm{aud}}$ to $[0,1]$ within environment, it computes each history's action-value \emph{range}, $\max_a Q^{\mathrm{aud}}(h,a)-\min_a Q^{\mathrm{aud}}(h,a)$, and retains the top quartile.
Its IQD--planning-error correlation is $\rho=.75\ [.68,.81]$.
This range is not the top-one-versus-top-two decision margin and is not interpreted as such.

\subsection{Spurious-coordinate validation}

Using training histories only, we select 500 coordinates from 5,000 cached trajectory statistics derived from the existing 100-test library; this control requires no additional simulator rollouts.
Selection requires $|\rho(\xi,\max_a Q^{\mathrm{aud}}_a)|<.05$ and $|\rho(\xi,\mathrm{tag})|>.3$.
The tag is a deterministic function of an analysis-only synthetic episode seed that indexes cached trajectory-statistic construction but never modifies the agent's coordinate observation.
Selecting cached response-trajectory statistics for association with this tag injects a statistical relation, not an environment noise channel; neither tag nor seed affects task dynamics, reward, termination, or state.
On held-out validation histories, the selected set retains substantial aggregate tag association (median $|\rho|=\NuisanceTagMedian$; \NuisanceTagAbovePointThreePct{} remain above $.3$).
Against the complete action-value vector, the median and 95th percentile of $\max_a|\rho(\xi,Q^{\mathrm{aud}}_a)|$ are $\NuisanceActionValueMedian$ and $\NuisanceActionValueNinetyFifth$; \NuisanceActionValueBelowPointOnePct{} are below $.10$.
Thus the coordinates are largely orthogonal to the audited control signal rather than merely to its maximum.
For reproducibility, let $m_k,M_k$ be coordinate $k$'s extrema on training histories and define
\[
\bar\xi_k(h)=\operatorname{clip}\!\left((\xi_k(h)-m_k)/(M_k-m_k),0,1\right),
\quad \nu(h,h')=500^{-1}\!\sum_k|\bar\xi_k(h)-\bar\xi_k(h')|.
\]
The unrescaled control gives every coordinate the same weight as each of 100 useful components: $d_{\mathrm{append}}=(100d_{\widehat\Tests}+500\nu)/600$.
For the equal-component-energy control, under the fixed response-stratified training-pair distribution $\mu$, set
\[
r=\sqrt{\E_\mu[d_{\widehat\Tests}^2]/\E_\mu[\nu^2]},
\qquad d_{\mathrm{eq}}=\tfrac12(d_{\widehat\Tests}+r\nu).
\]
Thus the useful component $d_{\widehat\Tests}$ and spurious component $r\nu$ have equal component energy, defined as their individual second moments under $\mu$; no fixed fraction of squared mixture energy is claimed because it also contains a cross term.
The extrema and $r$ are fitted once on training histories and frozen.

\section{Implementation and Compute Accounting}
\label{app:compute}

\paragraph{Cached response profiles and pair sampling.}
Each environment supplies 10,000 cached training histories collected by a uniformly random deployment-action policy to provide broad reachable-state coverage.
Every history has 100 uniformly weighted tests ($w_\tau=.01$), each response estimated from 100 rollouts.
QP and MD use a separate cached-history auxiliary batch of 256 histories and all $\binom{256}{2}=32{,}640$ within-batch pairs.
The main sampler clusters response vectors into 20 $k$-means groups and samples equally across groups to balance near and far relations; MD uses the identical sampler.
The uniform-pair control samples histories without this stratification.
This 256-history auxiliary batch is distinct from the base RSSM batch of 50 replay sequences.

\paragraph{Scale implementation.}
For each QP batch, $\widehat c=(\sum d_{ij}d_{Z,ij})/(\sum d_{Z,ij}^2)$ is computed analytically with stop-gradient, then stabilized by a $.99$-decay moving average initialized at one and clipped to $[.1,10]$; a negligible denominator skips the update.

\paragraph{Shared model and optimization.}
The 256-dimensional RSSM state comprises a 200-dimensional GRU component and a 56-dimensional stochastic-posterior mean; IQD uses both.
All methods use Adam at $3\times10^{-4}$ with $(\beta_1,\beta_2)=(.9,.999)$, 200,000 steps, a $10^6$-transition buffer, and discount $.99$.
CEM uses horizon 10, population 200, 10 iterations, and 20 elites.
There are 10 training seeds per condition and 2,000 held-out IQD pairs.

\begin{table}[ht]
\floatconts
  {tab:lossweights}
  {\caption{Global coefficient selection. Each candidate receives five runs per domain on tuning seeds 0--4; the equal-weight fixed-suite ID criterion selects one coefficient per method for every domain and terminal seed.}}
  {\centering\footnotesize\setlength{\tabcolsep}{7pt}
  \begin{tabular}{lll}
  \toprule
  \bfseries Method & \bfseries Candidate coefficients & \bfseries Selected coefficient\\
  \midrule
  TR & $\TuningCandidates$ & $\TuningTRSelected$\\
  MD & $\TuningCandidates$ & $\TuningMDSelected$\\
  BD-S & $\TuningCandidates$ & $\TuningBDSSelected$\\
  QP & $\TuningCandidates$ & $\TuningQPSelected$\\
  \bottomrule
  \end{tabular}}
\end{table}

Across four methods this entails \TuningRunsTotal{} tuning runs; no shift or final-evaluation episode enters selection, and evaluation intervals condition on the choices.
RNC's neighborhood threshold $.2$ and temperature $.1$ are fixed a priori and are not swept.
Terminal-model seeds 10--19 are disjoint from tuning seeds 0--4.

\paragraph{Target and model compute.}
Posterior construction and the response build each cost 112 CPU-core-hours per environment (672 each across the suite); the latter contains $10{,}000\times100\times100$ rollouts, approximately $5\times10^8$ simulator steps at mean continuation length five.
Training-time target/posterior precomputation totals 1,344 CPU-core-hours without GPU use, approximately seven wall-clock hours per environment on 32 cores when the two stages are run serially.
This figure excludes held-out response recomputation, evaluation, and calibration-analysis compute.

\begin{table}[ht]
\floatconts
  {tab:trainingcompute}
  {\caption{Audited A100 terminal-model training compute over six domains and \ComputeTerminalModelsPerMethod{} models per method (\ComputeTerminalModelsTotal{} total); hyperparameter-sweep training is excluded.}}
  {\centering\footnotesize\setlength{\tabcolsep}{8pt}
  \begin{tabular}{lccccc}
  \toprule
  \bfseries Method & RNC & TR & MD & BD-S & QP\\
  \midrule
  GPU-hours/seed & $4.2$ & $4.1$ & $4.4$ & $4.0$ & $4.5$\\
  Total GPU-hours & $252$ & $246$ & $264$ & $240$ & $270$\\
  \bottomrule
  \end{tabular}}
\end{table}

The five totals sum to \ComputeTerminalGPUHours{} GPU-hours for selected terminal models.
The \TuningRunsTotal{} tuning jobs cost approximately \ComputeTuningGPUHours{} GPU-hours, robustness controls and architecture checks approximately \ComputeRobustnessGPUHours{}, and the frozen visual encoder \ComputeVisualEncoderGPUHours{}.
Including all reported learned models and the encoder gives approximately $\ComputeTotalGPUHours{}$ A100 GPU-hours.

\begingroup
\footnotesize
\setlength{\parskip}{0pt}
\section{Additional Experimental Protocol}
\label{app:protocol}
\enlargethispage{2\baselineskip}

\paragraph{Test-response regression (TR).}\leavevmode\\
TR uses a $256\!\to\!128\!\to\!d_{\mathrm{out}}$ ReLU MLP on the RSSM posterior-mean state; categorical outcomes use cross-entropy.
Each CartPole test predicts a five-component Gaussian mixture and minimizes sample-based one-dimensional Wasserstein distance between 1,000 predicted samples and 100 empirical returns.
It uses \citet{blondel2020sorting}'s $\ell_2$-regularized soft-sort (strength $1.0$) and soft Gumbel--Softmax assignments (temperature $.5$).
Output dimensions across 100 tests are 400 (both GridWorlds), 200 (Key--Box/MiniWorld), 300 (MiniGrid), and 1,500 (CartPole).

\paragraph{Metric decoding (MD).}
MD applies a 262,913-parameter $768\!\to\!256\!\to\!256\!\to\!1$ ReLU MLP to $[z_i+z_j;z_i\odot z_j;|z_i-z_j|]$ and minimizes squared error to $d_\Tests(h_i,h_j)$ without constraining native distance.

\paragraph{Belief and privileged-state objectives.}
BD-S applies a $256\!\to\!128$ ReLU trunk to Appendix~\ref{app:estimation}'s posterior targets; continuous heads use Gaussian/mixture likelihoods with softplus scales and bivariate tanh correlation, while categorical heads use cross-entropy.
PH uses the same hidden layers but squared-error predicts the realized 2D GridWorld goal, making it an oracle auxiliary control.

\paragraph{Response-neighborhood contrastive (RNC).}
RNC uses $d_\Tests<.2$ positives in multi-positive InfoNCE with similarity $-\|z_i-z_j\|_2^2/.1$ on raw posterior-mean states, no projection head or hard-negative mining, and ignores anchors without positives.

\paragraph{Quotient preservation (QP).}
QP uses Equation~\ref{eq:qploss}, stop-gradient on the analytic batch-optimal scale; it aligns distances but does not directly minimize held-out normalized IQD.

\paragraph{Additional protocol details.}
Appendices~\ref{app:shiftaudit} and~\ref{app:ladder} audit leakage/taxonomy and specify the bootstrap.
For the memory audit, a held-out history follows its recorded actions through $\Delta\in\{1,\ldots,20\}$ learned transitions without observations; a disjoint-episode linear goal probe and the fixed horizon-10 CEM planner measure recoverability and terminal-choice error, respectively.
Thus $\Delta$ is rollout depth, not planning horizon.
The missing-probe condition removes the goal-revealing test.
Outcome noise independently corrupts 20\% of response outcomes.
The spurious-coordinate condition appends the 500 training-selected coordinates validated in Appendix~\ref{app:coverage}; the equal-component-energy condition uses the frozen construction defined there.
Shuffled stress permutes within-batch targets, preserving their marginal; uniform sampling removes response clustering, and mean-only TR discards higher distributional information.
The MiniGrid check uses TD-MPC-style latent dynamics/value learning without reconstruction; Value-only, RNC, and QP share planner, capacity, transitions, and privileged target budget (Table~\ref{tab:architecture}).
\endgroup

\end{document}